# StreamSoNG: A Soft Streaming Classification Approach

Wenlong Wu, *Student Member*, *IEEE*, James M. Keller, *Life Fellow*, *IEEE*,
Jeffrey Dale, *Student Member*, *IEEE*, and James C. Bezdek, *Life Fellow*, *IEEE*

*Abstract*—Examining most streaming clustering algorithms leads to the understanding that they are actually incremental classification models. They model existing and newly discovered structures via summary information that we call footprints. Incoming data is normally assigned a crisp label (into one of the structures) and that structure's footprint is incrementally updated. There is no reason that these assignments need to be crisp. In this paper, we propose a new streaming classification algorithm that uses Neural Gas prototypes as footprints and produces a possibilistic label vector (of typicalities) for each incoming vector. These typicalities are generated by a modified possibilistic k-nearest neighbor algorithm. The approach is tested on synthetic and real image datasets. We compare our approach to three other streaming classifiers based on the Adaptive Random Forest, Very Fast Decision Rules, and the DenStream algorithm with excellent results.

## I. INTRODUCTION

Consider the following problem, An autonomous mobile agent has to locate particular objects over a large, and potentially unknown area. This agent could be a vehicle or flying drone equipped with a variety of Electro-Optical/Infra-Red (EO/IR) and depth sensors, or could be an underwater drone, an Unmanned Underwater Vehicle (UUV) using Synthetic Aperture Sonar (SAS). Let's pick the latter as an example, and assume we wish to detect sunken pirate booty on the seafloor. The objects of interest will present different signatures based on the surrounding seafloor environment such as bare sand, sea grass, sand ripples, coral, mud, rocks, etc. Some backgrounds will allow for simple classifiers, whereas others require more sophisticated algorithms. But the key is for the UUV to recognize the current environment to determine the correct tailored object classifier. More realistically, the current SAS image can easily be a blend of multiple seafloor textures, including as yet unseen environments, and so a blend of classifiers will provide the best chance at detecting the treasure. This example, derived from an actual ongoing project, illustrates the need for streaming classification that can handle diverse and changing class definitions (e.g., seafloor textures) while being able to identify new, but unknown environments. The streaming classifier needs to be able to provide soft (fuzzy, probabilistic, or possibilistic) labels that can be used for object classifier fusion. The focus of this paper is that initial component: streaming soft classification to create a possibilistic label vector to be fed to a secondary object detection system.

Data stream processing techniques have gained much attention in recent years. Streaming data, such as social network clique information or daily sensor firing information, are generated every day. Conventional clustering and classification models use static (batch) data and hence, are not directly applicable to data streams. Therefore, alternate strategies are required to incrementally update models as new feature vectors become available. In [1], the authors argue that in fact, all streaming data analysis should be thought of as classification. There has been some research on classification in data streams [2], such as adapting CluStream as an on-line classifier [3], Very Fast Decision Trees [4], [5], rule based classifiers [6], and a nearest neighbor technique [7]. However, for the most part, these approaches are collectively referred to as streaming clustering, and there are a number of ways to organize them into taxonomies [8], [9].

The multitude of streaming data analysis algorithms have several things in common. First, they do not retain the entire dataset. They maintain "footprints" that summarize the structures discovered, and they usually have a mechanism to incrementally update footprints as new vectors arrive. In [10]–[14], underlying probabilistic models are used and the footprint contains probability distribution parameters. Many of the density based streaming models also contain footprint entries that allow calculation of basic summary statistics [15]–[19]. The footprints can additionally contain the structure of fuzzy rules [20], [21]. In [22], an evolving system was proposed that could add, remove, merge and split clusters. These streaming methodologies generally maintain only summary information, i.e., footprints, and a means to incrementally update it. Some of the streaming methods are used in different real world applications such as identification, control and fault detection [23]–[27].

The other basic trait of these algorithms is that they assign labels to each point as it appears. For the most part, the labels are crisp – a point is assigned completely to an existing or new structure. Density-based algorithms put the point into a micro-cluster and underlying probability models, like Gaussian mixtures, use maximum likelihood to make the assignment. Once the points are assigned to a structure, their labels cannot change, as only the footprints are preserved. Footprints have also been called summary information (prototypes, statistical properties, etc.) which are used to decide membership or the typicality of incoming data points, and whether a new class/cluster is needed to describe the data. In fact, crisp labels are desirable because each incoming point is used to update the appropriate footprint. There is no iteration over the actual data, though some algorithms do form larger clusters by offline clustering of multiple footprints. Hence, what we call streaming clustering is actually classification, and should be thought of in that way.

Creating a crisp label vector implies that each point is assigned to one and only one existing structure. The newly labeled point is used to update the structure footprint. The exception to this rule is when an incoming sample is judged by the algorithm to be an outlier with respect to previously seen data. The point doesn't receive a current class label, nor is it used to update the footprint. Different approaches deal with outliers in different ways, but they are the keys to discovery of new structure in the stream.

We note that, while not specifically related to streaming data analytics, a related vein of research is called open set recognition [28], in which a classifier can recognize a closed set of objects but must operate on an open test set. This means that the classifier must have the ability to identify when samples do not belong to any class upon which it was trained. Approaches to achieving open set recognition include leveraging a series of one-vs-all support vector machines (SVMs) [28], the Nearest Non-Outlier (NNO) algorithm [29], and, more recently, the OpenMax model for deep neural networks [30].

The motivating streaming algorithm that we use in this paper is a variant of the Missouri University (MU) streaming clustering algorithm called MUSC. The version, MUSC II [12], is a modified form of the original method that is found in [10] and [11], where in both versions, the underlying footprints are modeled by the parameters of the components of a Gaussian Mixture. In MUSC II, initialization, and the search for new structures, is done using the *sequential possibilistic one-means with dynamic eta* (SP1M-DE) algorithm [31]. In that variant, crisp label vectors (including an "outlier" label) are assigned to each incoming point. The outliers are saved and examined frequently to detect new structures.

There is no fundamental reason that crisp label vectors need to be assigned to incoming data. Our initial motivating example shows the desirability of soft labels in an object detection system where different classifiers are used in different contexts. In that scenario, a UUV traveling over different environments may employ different classifiers for detecting objects based on the observed seafloor type, like sand ripple, seagrass, etc. A baseline application of this type is described in [32], but without a streaming environment module. The environment-detecting component can be initialized with suspected known backgrounds. As new environment imagery is acquired, it must be matched to the known backgrounds to determine the appropriate classifier to use. However, the environment in a particular SAS image isn't always an exact match to the known backgrounds; it can either be partially representative of more than one background, or something completely new. This is where possibilistic labels are particularly valuable.

To be precise, if $X = \{x_1, \cdots, x_n\}$ represents a set of objects, or a set of feature vectors for the objects, then a collection of subsets and membership functions $\{(A_1, \mu_1), \cdots, (A_C, \mu_C)\}$ is called a

1. Crisp partition if $\mu_i: X \to \{0,1\}$ and

    $\sum_{i=1}^{C} \mu_i(x_k) = 1 \; for \; k = 1, \cdots, n.$

2. Fuzzy or Probabilistic partition if $\mu_i: X \to [0,1]$ and $\sum_{i=1}^{C} \mu_i(x_k) = 1 \; for \; k = 1, \cdots, n.$

3. Possibilistic partition if $\mu_i: X \to [0,1]$ and

    $0 < \sum_{k=1}^{n} \mu_i(x_k) \le n \; for \; i = 1, \cdots, C$, and

    $\max_i \mu_i(x_k) > 0 \; for \; k = 1, \cdots, n$.

The vector $(\mu_1(x_k), \cdots, \mu_C(x_k))$ of class membership values is called the label vector for $x_k$. For a crisp label vector, $x_k$ belongs to one and only one of the subsets; for a fuzzy label vector, the membership of $x_k$ is shared among the subsets (though constrained to sum to 1); but for a possibilistic label vector, the membership of $x_k$ is unconstrained and reflects the typicality of $x_k$ to each of the subsets. Hence, in a possibilistic framework, a particular patch of seafloor can have a high typicality to more than one known seafloor type (when there is a blend) or it can have a very low typicality to all of the currently known seafloor types, signaling an outlier or new seafloor type.

If a streaming vector (new image) can be assigned typicalities in the known backgrounds, then classifiers can be blended. Images with low typicality in all known backgrounds are listed as outliers and some standard classifier fusion can be used. When a new structure (background class) is found in the data stream, either an automatically generated label can be assigned, or an active learning phase can be initiated in which a human will need to interact with the system to give the new class a label and perhaps then have a new classifier inserted.

We are mainly interested in the environment recognition portion of such a system. We propose a new streaming classification algorithm called *streaming soft neural gas* (StreamSoNG). The StreamSoNG algorithm uses the neural gas algorithm (NG) [33] during initialization to find sparse data representations for known classes, i.e., to generate the class footprints. A modified version of the possibilistic k-nearest neighbors algorithm (PKNN) [34] is employed as the streaming classifier to assign soft (possibilistic) labels to data stream vectors. The typicality vector of possibilistic labels can be used directly in a classification scheme. Based on maximum typicality, class footprints are incrementally updated. This updating allows for class definitions to vary from their initial topologies. An incoming sample that has low typicality values in all currently known classes is marked as an outlier and saved to an anomaly list. The sequential possibilistic one-means algorithm (SP1M) [31] is run the anomaly list to identify a potential new class.

The rest of the paper is organized as follows. Section II briefly reviews NG, PKNN and SP1M algorithms. Section III introduces our StreamSoNG algorithm. Section IV describes the four synthetic datasets used in this paper. Section V shows our experimental results. Section VI summarizes our conclusions and future work.

## II. BACKGROUND

### A. Neural Gas

The *neural gas algorithm* (NG) [33] is a competitive-learning neural network algorithm in the same family as the self-organizing feature map algorithm (SOFM) [35]. The NG algorithm aims to optimally describe the topology of data vectors using a fixed number of prototypes.

In the standard NG, given a set of vectors $X = \{x_t | t \in \mathbb{N}\}, X \subset R^q$ and a finite number of prototype vectors $p_i, i = 1, \dots, N, p_i \subset R^q$, a data vector $x_t$ at timestamp $t$ is randomly chosen from $X$. The distance order of the prototype vectors to the chosen data vector $x_t$ is computed using a chosen measure of distance on $R^q$. Let $i_1$ denote the index of the closest prototype vector, $i_2$ denote the index of the second closet prototype vector, and $i_N$ denote the index of the prototype vector most distant to $x_t$. Then each prototype vector is adapted according to

$$p_{i_k}^{t+1} = p_{i_k}^t + \varepsilon e^{-\frac{k}{\lambda}}(x_t - p_{i_k}^t), k = 1, \dots, N \quad (1)$$

where $\varepsilon$ is the adaptation step size and $\lambda$ is a neighborhood range parameter. Both $\varepsilon$ and $\lambda$ are reduced according to a predefined schedule with increasing iterations. After sufficiently many epochs using randomly sampled data vectors, the prototype vectors cover the data space with minimum representation error, i.e., the sum of squared errors of each point to its closest prototype.

### B. Possibilistic K-Nearest Neighbors

The PKNN algorithm extends the crisp KNN algorithm that first assigns a fuzzy membership (between 0 and 1) to each training pattern rather than using a binary class membership. The membership is assigned as described in [36] using

$$\mu^i(p) = \begin{cases} 0.51 + \left(\frac{n_i}{K}\right) \times 0.49, & \text{if } i = j \\ \left(\frac{n_i}{K}\right) \times 0.49, & \text{if } i \neq j \end{cases} \quad (2)$$

where $n_i$ denotes the number of neighbors that belong to the $i^{th}$ class, i.e., $\sum_{i=1}^C n_i = K$ and $j$ is the actual class label of protype $p$. Note, for StreamSoNG, we are assigning training vector memberships to the NG prototypes that make up class footprints. The pattern's fuzzy membership $\mu^i(p)$ controls its contribution during the classification process.

The version of PKNN proposed in [34] assigns membership values (typicalities) $t^i(x)$ of a data vector $x$ to class $i$ using

$$t^i(x) = \sum_{k=1}^K \mu^i(p_k) w(x, p_k) \quad (3)$$

where $p_k$ is the $k^{th}$ nearest prototype to $x$ and

$$w(x, p_k) = \frac{1}{1 + \left[\max\left(0, \|x - p_k\| - \frac{\eta_1}{\eta_2}\right)\right]^{\frac{2}{m-1}}} \quad (4)$$

In Equation (4), $\eta_1$ and $\eta_2$ are constants that are estimated from the training data and $m > 1$ is a "fuzzifier" parameter. One approach is to identify the five nearest prototypes to each training sample and construct a histogram containing all associated distances. Then, we take $\eta_1 = \mu_H$ and $\eta_2 = 3 \times \sigma_H$, where $\mu_H$ and $\sigma_H$ are the mean and standard deviation of the histogram of distances [34].

The expression in Equation (4), worked well for a two-class application. In the streaming scenario, the number of classes is unknown. We will formulate a version more closely aligned to the original PCM, the updates as shown in equations (5) and (6) that naturally extends to any number of classes.

### C. Sequential Possibilistic One-Means

The *sequential possibilistic one-means algorithm* (SP1M) was developed from the *possibilistic C-means* (PCM) clustering algorithm [37]–[39] for static data. In static PCM, let $X = \{x_1, \cdots, x_N\}$, $X \subset R^q$ and $C$ = the number of clusters in $X$. Each cluster is independent of the others and, effectively, is found separately by iterating between the following equations:

$$u_{ij} = \frac{1}{1 + \left(\frac{d_{ij}^2}{\eta_i}\right)^{\frac{1}{m-1}}} \quad (5)$$

$$v_i = \frac{\sum_{j=1}^N u_{ij}^m x_j}{\sum_{j=1}^N u_{ij}^m} \quad (6)$$

where $u_{ij}$ is the typicality value of $j^{th}$ data vector to the $i^{th}$ cluster center $v_i$, for $j = 1,\dots, N$, and $i = 1, \dots, C$, $d_{ij}$ is the Euclidean distance value of $j^{th}$ data vector to the $i^{th}$ cluster center $v_i$, $\eta_i$ (a cluster-based hyperparameter) is the distance from the cluster center at which the typicality value in that cluster is equal to 0.5, and $m > 1$ is the fuzzifier value. See [26] for discussions about these parameters.

The family of SP1M algorithms [31], [37], [38] sequentially search for one cluster at a time and choose the starting point probabilistically at a rate inversely proportional to the maximum typicality of each point with respect to the currently discovered clusters. These variants reject centroids that are coincident with one of the previously discovered prototypes.

The SP1M pseudocode is shown in Table I. For additional details about the SP1M algorithm, see [31].

TABLE I: SP1M PSEUDO CODE

| |
|---|
| Input: *X*, *C*, *ε* |
| Output: *U*: Final membership partition |
|         *V*: Final cluster centers |
| 01: Initialize *U*, *V* as empty |
| 02: Do { |
| 03: ---- Repeat <loop to find a suitable cluster> |
| 04: ---- ---- Pick $v \in X$ from probabilities |
| 05: ---- ---- Repeat <loop to execute P1M> |
| 06: ---- ---- ---- Compute $\eta$ dynamically (*) |
| 07: ---- ---- ---- Compute *u* (*v*, *X*)   (5) |
| 08: ---- ---- ---- Compute *v* (*u*, *X*)   (6) |
| 09: ---- ---- Until cluster center is stable |
| 10: ---- Until no coincident cluster is found |
| 11: ---- Append *u* to U |
| 12: ---- Append *v* to V |
| 13: } While (i + +< *C* && #(P1M) < *K*) |

*details of dynamic $\eta$ computation, along with the initialization probabilities and the stopping criterion are discussed in [31]

## III. STREAMING SOFT NEURAL GAS

In this paper, we propose a new algorithm called the *streaming soft neural gas algorithm* (StreamSoNG) to classify streaming data vectors. The novel aspects of this technique include using NG prototypes as class footprints, a different PKNN formulation both in the initialization phase and in the streaming portion of a possibilistic label assignment for incoming features, and in the incremental update of class footprints. During initialization, we use NG to learn representations (prototypes) of the initial $C$ classes, $\{p_{ij} | i = 1, \dots, C; j = 1, \dots, n_i\}$, and only save the learned prototypes along with their labels as the class footprints.

We compute each prototype's distance to its $K$ nearest prototypes (here, $K$ is 5) in each existing class, then build a histogram of these distances for each class as shown in Fig. 1 (a). The region in feature space that each prototype's region of influence (the parameter $\eta_i$ in Equation (5)) is estimated as the mean of the distance histogram using Equation (7).

$$\eta_i = \frac{1}{n * K} \sum_{j=1}^{n} \sum_{k=1}^{K} \|p_{kj} - p_j\| \quad (7)$$

where $n$ is the number of neurons in each class, $K$ is the number of nearest neighbors, $i$ is the cluster number index.

Fig. 1 (b) shows the value of $\eta$ plotted on the initialization set of the dataset 1. The $\eta_i$ value in a class is the radius of the small circles in Fig. 1 (b). As we see, each prototype covers a potentially overlapping sub-area of the data distribution in each class. The clusters in the example dataset in Fig. 1 (b) were created by equal size circular Gaussians, so the values of $\eta_i$ turn out to be the same for each class in this example. However, that need not be the case for more complex class definitions.

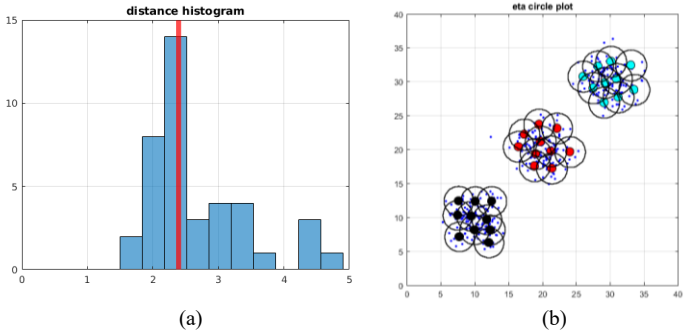

Fig. 1. (a) Distance histogram in one class (b) $\eta$ circle plot on the initialization set of dataset 1

When streaming data $x_t$ arrives, we first get its $K$ nearest prototypes and compute the prototypes' fuzzy label memberships using Equation (2) and the typicalities of streaming data to its $K$ nearest prototypes using Equation (8) below. The typicality value $t_{ik}(x_t)$ of the streaming data $x_t$ to its $k^{th}$ nearest prototype is computed as in the original PCM using

$$t_{ik}(x_t) = \frac{1}{1 + \left(\frac{\|x_t - p_{ik}\|^2}{\eta_i}\right)^{\frac{1}{m-1}}} \quad (8)$$

Here, $i$ is the class label for the $k^{th}$ nearest prototype, and $\eta_i$ is estimated from the histogram of the distance between prototypes within a class, as shown in Fig. 1. Normally, and for the experiments in this paper, the Euclidean norm is used. Then, we multiply the fuzzy label memberships and typicalities to get the scaled typicalities with fuzzy labels, $t'_{ik}(x_t)$, using Equation (9).

$$t'_{ik}(x_t) = \mu^i(p_{ik}) * t_{ik}(x_t) \quad (9)$$

In other words, we compute the scaled typicality $t'_{ik}(x_t)$ to each of the $k$ prototypes using the typicality from Equation (8) and the fuzzy labels $\mu^i(p_{ik})$ from Equation (2) for the prototypes of each class to get $t'_{ik}(x_t)$, and then compute the class average typicality using

$$\bar{t}_i = \frac{1}{K} \sum_{k=1}^{K} t'_{ik}(x_t) \quad (10)$$

and pass them to a scaling function, Equation (11) as its class typicality.

$$T_i(x_t) = \begin{cases} 0, & \bar{t}_i \leq 0 \\ 2 * \bar{t}_i - \bar{t}_i^2, & 0 < \bar{t}_i \leq 1 \\ 1, & \bar{t}_i > 1 \end{cases} \quad (11)$$

Now we have the class typicality vector, $T(x_t) = (T_1(x_t), \cdots, T_C(x_t))^T$ of the streaming data $x_t$ and use the maximum class typicality to represent the typicality of the streaming data to its closest class. If the maximum class typicality value is larger than a preset threshold, we assign the label of its closest class to this streaming data for footprint update. At that time, we update the prototypes that are in the same class of the current streaming data point according to

$$p_{ik}^{t+1} = p_{ik}^t + \alpha * T_i(x_t) * e^{-\frac{k}{\lambda}}(x_t - p_{ik}^t) \quad (12)$$

where $\alpha$ is a learning rate (we use 0.1 in this paper); $p_{ik}^t$ is the $k^{th}$ closest prototype (neuron) to data vector $x_t$ at time $t$; $\lambda$ is a neighborhood range parameter (we use 2 in this paper). The typicality value $t'_{ik}(x_t)$ measures the typicality of a streaming data vector $x_t$ to the prototype $p_{ik}^t$. If $x_t$ has a high typicality to a given neighbor prototype, meaning that it is a good representation of that class, then we update the $k^{th}$ nearest prototype with a large step; otherwise, we only update the $k^{th}$ nearest prototype by a small amount.

If the maximum class typicality value is smaller than a preset threshold, then the streaming data point has low connection to any class. In this case, we mark the streaming data as an unseen class (outlier) and save it to the outlier list $O$ for future analysis.

If the streaming data point is marked as an outlier, we run P1M on the updated outliers list $O$ to search for a new class. If P1M finds a cluster for which the number of points with typicality bigger than 0.5 is larger than a minimum cluster-formed threshold, we identify this subset as a new class, run NG on it, and remove these points from the outlier list $O$. The newly generated prototypes will be appended to the current learned prototypes and represent the new class. At this point, or actually at any time there is an outlier, the system can ask a human to provide a semantic class label or can reject an outlier completely.

The pseudocode of the streaming soft neural gas algorithm is shown in Table II.

TABLE II. STREAMING SOFT NEURAL GAS PSEUDOCODE

**Initialization**
Input: initialization set $X\_init$ and class label $y\_init$;
Output: prototypes $P$;
01: for $i$ in each class $y\_init$:
02: ---- run NG on class($i$) in $X\_init$;
03: ---- save neurons of each class($i$) into prototypes $p_{ik}, k = 1, \ldots, n_i$;
04: end for

**Stream Processing**
Input: streaming set $X$, initial prototypes $P$, typicality threshold $t$, minimum number of points $M$ to form a new class;
Output: streaming set class label vector $L$ and class typicality vector $T$
01: initial PKNN model with $P$ (declare PKNN);
02: for $x$ in streaming set $X$:
03: ---- compute $K$ nearest prototypes' fuzzy label memberships using Eq. (2)
04: ---- compute typicalities $t_{ik}(x)$ of $x$ to its $K$ nearest prototypes in $P$ using Eq. (8)
05: ---- multiply typicalities with fuzzy label memberships to get the scaled typicalities $t'_{ik}(x)$ using Eq. (9)
06: ---- compute the class typicalities of $x$ by taking the average of the scaled typicalities $t'_{ik}(x)$ in each class
        using Eq. (10) and apply scaling function using Eq. (11)
07: ---- predict class label $L^i(x)$ and class typicality $T^i(x)$ using the largest class typicality
08: ---- if (class typicality $T^i(x) > t$):
09: ---- ---- update $P$ for class $i$ with $x$ incrementally using Eq. (12);
10: ---- else:
11: ---- ---- mark $x$ as an outlier and save to outliers list $O$;
12: ---- ---- run P1M on $O$ to search for a new cluster $C'$;
13: ---- ---- if (# of points with typicality>0.5 in $C' > M$):
14: ---- ---- ---- run NG on $C'$, and add the new prototypes to $P$;
15: ---- ---- ---- remove the points with typicality>0.5 in $C'$ out of $O$ and reset the outlier label in class label vector $L$ with the new class;
16: ---- ---- end if
17: ---- end if
18: end for

## IV. SYNTHETIC DATASETS

To test the StreamSoNG algorithm, we use four synthetic datasets and one real world data set. The first three synthetic data sets use Gaussian clouds to provide a structured and well understood environment. In the first dataset, the mean values of three Gaussian classes in the initialization set are (10, 10), (20, 20), and (30, 30). The mean values of two unknown (Gaussian) classes in the streaming set are (40, 40) and (50, 50). The covariance matrix in both the initialization and streaming sets are [4, 0; 0, 4]. In the second dataset, the mean values of three classes in the initialization set are (10, 10), (20, 20), and (30, 30). The mean values of two unknown classes in the streaming set are (40, 40) and (50, 50). The covariance matrix in both the initialization and the streaming set is [15, 0; 0, 15]. In the third dataset, the mean values of three initialization classes are (10, 20), (20, 30), and (30, 20), whereas the mean values of two unknown classes in the streaming set are (20, 10) and (20, 20). The covariance matrix in both the initialization and the streaming set is [5, 0; 0, 5]. The fourth dataset provides a non-Gaussian example with two "circular" initialization sets with "centers" at (10, 20), and (20, 15). The radius of the circles is around 10. The mean value of a new unknown (Gaussian) class in the streaming set is (40, 30) with a covariance matrix [10, 0; 0, 10]. All four synthetic datasets are two dimensional and the scatter plot of the four datasets is shown in Fig. 2. The top four scatter plots, (a) – (d), represent the initialization sets and the bottom four plots, (e) – (h), show the temporal sequence of the stream after initialization. The arrows in streaming sets show how the streaming data evolves over time.

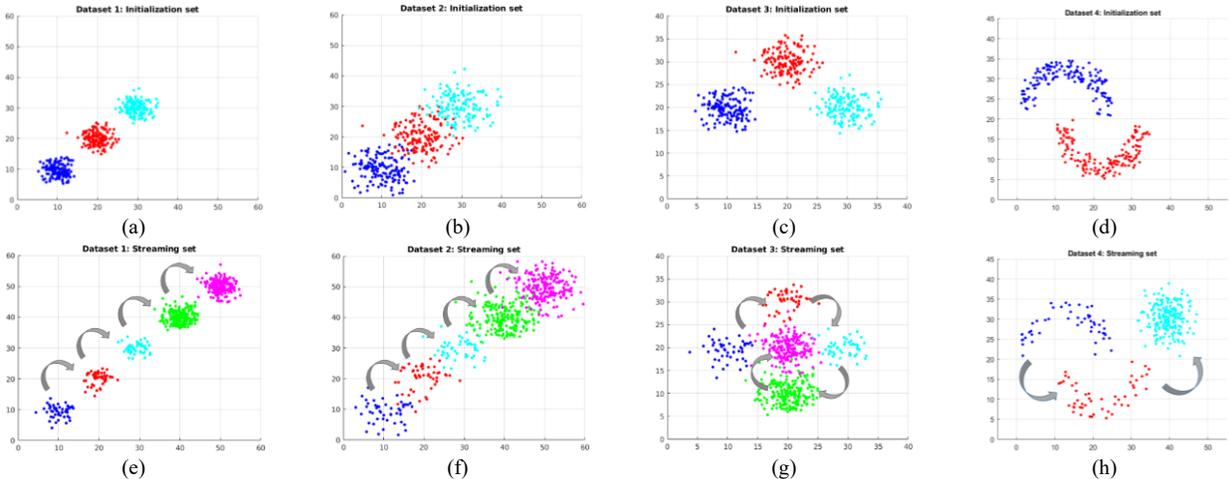

Fig. 2. The scatter plot of four synthetic datasets (class initialization data: figures (a) – (d) and streaming sets: figures (e) - (h))

## V. EXPERIMENTS

In this section, we run four experiments to test the StreamSoNG algorithm. The first experiment compares different neuron (prototype) update mechanisms on the four synthetic datasets. The second experiment studies the effect of permuting the presentation order of streaming data on the algorithm. The third experiment visualizes how the maximum typicality value of a specific data sample changes as the model updates with streaming data. The last experiment tests StreamSoNG algorithm on a real-world texture image dataset. We use the precision score to evaluate the models. The precision score compares the prediction with the ground truth, and is defined as

$$precision(A, B) = \frac{\sum_{i=1}^{N}(A_i = B_i)}{N} \quad (12)$$

where A is streaming prediction set, B is streaming ground truth set, and N is the samples number of the streaming set. The precision score is computed on the whole streaming set. Note that this measure is computed after hardening the possibilistic labels assigned to each streaming input vector. This makes our soft streaming classifier comparable to other crisp models.

### A. Parameter setting

There are some user-specified choices in Table II that need to be made for implementation of StreamSoNG. We used the number of neurons in each class $n = 10$, typicality threshold $t = 0.1$, minimum number of points to form a new class $M = 30$, epsilon $\varepsilon = 0.01$, fuzzifier $m = 1.5$, learning rate $\alpha = 0.1$ (in Equation (12)), neighborhood range lambda $\lambda = 2$ (in Equation (12)). We do not have room in this paper to perform experiments that establish constraints, rules of thumb, and recommendations in this paper, but we will make this a priority for a follow up study of the StreamSoNG algorithm.

### B. Experiment 1: Comparison of different neuron update mechanisms

Three neuron updating mechanisms can be used in data stream processing. The first one is to save all data samples and rerun the NG algorithm on the updated data samples to get new data representations (neurons). The second method is to update only the $K$ nearest neurons using Equation (12). The last method is to rerun the NG algorithm on the prototypes and the new streaming data sample, using it, in effect, as a potential prototype. Fig. 3 shows the precision scores of the three neuron updating mechanisms on the four synthetic datasets. The red dotted line is the first update mechanism that saves all data samples and reruns the NG algorithm on the entire updated samples. The blue line is the second update mechanism that updates the $K$ nearest neurons. The black dotted line is the third update mechanism that reruns the NG algorithm on the neurons and the new streaming data sample.

As we see in Fig. 3, the first method has the highest precision score but it requires more computation and data storage because it saves all data samples in both initialization and streaming sets. This method represents an upper bound but goes against the spirit of streaming data processing. The second method that updates the $K$ nearest neurons performs well compared to the first approach and clearly outperforms the third method that reruns the NG algorithm on the neurons and new streaming data sample. The third neuron updating mechanism can easily forget the learned representations. Therefore, updating $K$ nearest prototypes with streaming data is an accurate and efficient method to incrementally adjust the prototypes in a class.

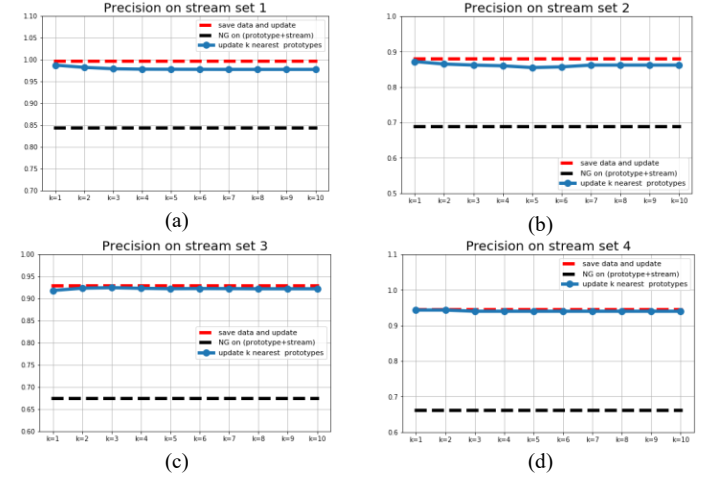

Fig. 3. Precision score on the four datasets with different neuron updating mechanisms

Furthermore, the StreamSoNG algorithm can not only produce a class label for a data sample, but also a typicality matrix that measures how well a data sample belongs to a specific class. If the typicality value of a streaming data sample is high, the algorithm is more confident about its prediction. In this experiment, we compute the precision scores for the predictions only where typicality values are higher than 0.2.

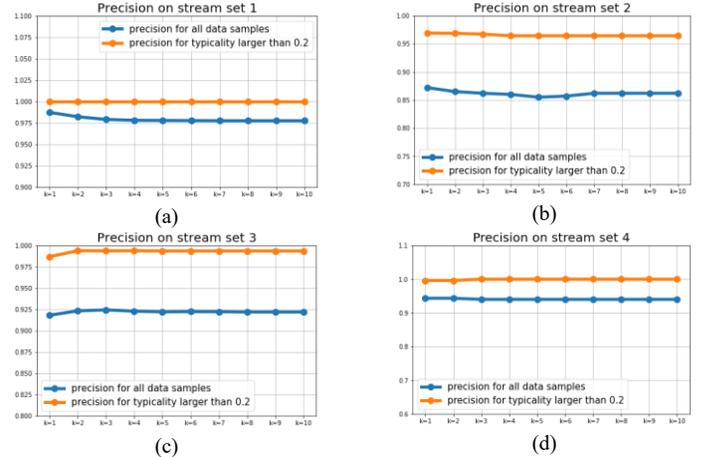

Fig. 4. Precision scores of high typicality samples on the four datasets with different values of $K$ in the PKNN

As we see in Fig. 4, the precision scores for typicality values higher than 0.2 are higher than the precision score for all streaming data samples. That is, the StreamSoNG algorithm has a higher precision score for its confident predictions.

In this experiment, we also run the *K-Nearest Neighbor* (KNN) algorithm [40], *Adaptive Random Forest* (ARF) classifier [41], *Very Fast Decision Rules* (VFDR) classifier [42], and DenStream algorithm [15] on the same synthetic datasets. The results are listed in Table III.

Table III. Precision scores of existing methods and StreamSoNG

|  | Dataset 1 | Dataset 2 | Dataset 3 | Dataset 4 |
|---|---|---|---|---|
| KNN(K=3) [40] | 0.271 | 0.259 | 0.271 | 0.331 |
| ARF [41] | 0.182 | 0.117 | 0.095 | 0.167 |
| VFDR [42] | 0.273 | 0.266 | 0.273 | 0.327 |
| DenStream [15] | **1** | 0.073 | 0.627 | 0.82 |
| StreamSoNG(K=3) | 0.979 | **0.862** | **0.924** | **0.94** |

As we can see in Table III, the KNN, ARF, and VFDR schemes have low overall precision scores because they cannot detect the new classes in data streams but only deal with the concept drift problem. The lack of new classes in evaluation causes their low scores. DenStream has a perfect score on the well separated dataset (dataset 1) but has a poor score on the overlapping cluster dataset (dataset 2) because it merges close structures. The StreamSoNG algorithm has the highest overall precision score because it can not only detect new classes but also works well on the overlapping cluster dataset.

### C. Experiment 2: the effect of permuting the streaming data

Suppose that the streaming data does not follow the specific pattern of arrival as we assumed in experiment 1. Here, we shuffle the order of streaming set and re-run the StreamSoNG algorithm on the shuffled streaming set. The precision scores using the k nearest neurons update mechanism on four synthetic datasets are shown in Fig. 5 as a function of k.

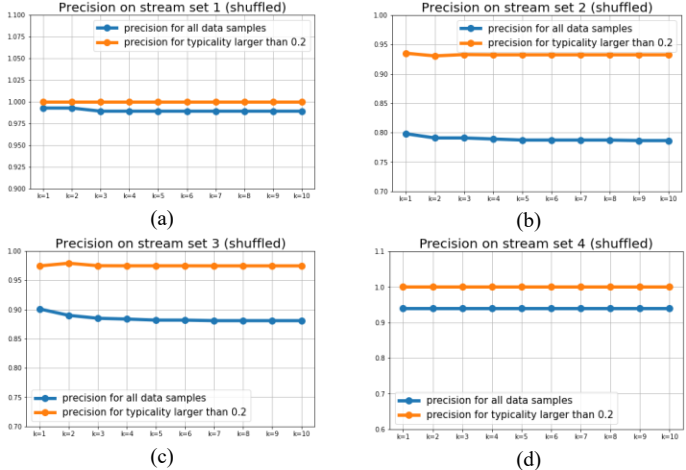

Fig. 5. Precision score on the four shuffled datasets with different values of k in the PKNN

As we see in Fig. 5, the precision scores on datasets 1 and 4 stay very close to the precision scores on the unshuffled streaming set because the classes in datasets 1 and 4 are well separated. The precision scores on datasets 2 and 3 decrease on the shuffled streaming sets compared to the precision scores on the unshuffled streaming set. This is because the two clusters in streaming set are very close to each other and it is hard to distinguish them at the beginning with randomly presented vectors. In addition, we compute the precision scores for the streaming samples with typicality values higher than 0.2. As before, the precision scores with confident predictions are higher than the precision score for all streaming data samples.

### D. Experiment 3: Visualization of typicality value changes in streaming data

In this experiment, we track the typicalities of several data samples and see how they change as the model updates with streaming data. It is as if these points are presented repeatedly, after each real sample of the data stream. They are not used to update the class footprints, but only to monitor changes in maximal typicality throughout the process.

Fig. 6 shows how the streaming data in dataset 1 and the maximum typicality value of four points evolve over time. Fig. 6 (a), (c), (e) plot the streaming data at time $t_1$, $t_2$, $t_3$ and 4 diamond symbols (in green, red, cyan, magenta color) in the data plots are studied. Fig. 6 (b), (d), (f) plots the maximum typicality value of the four points in different colors at the different time (the colors in one row matches).

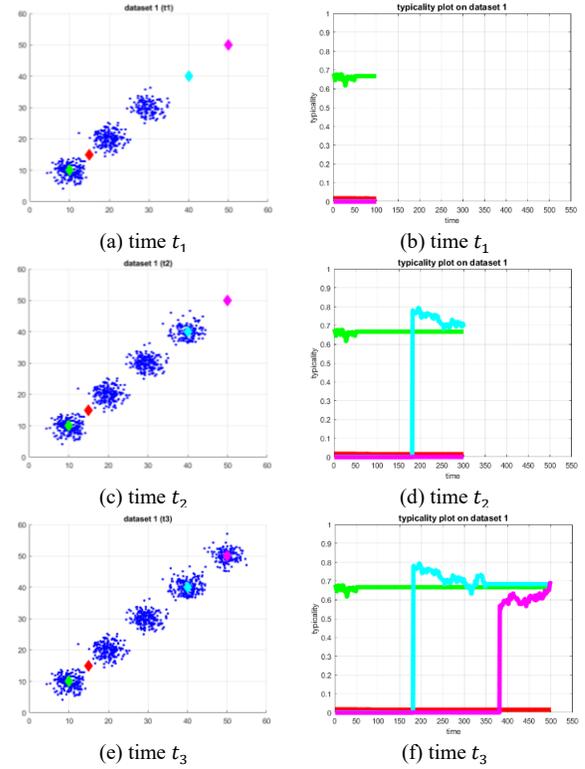

Fig. 6. (a), (c), (e) Streaming data plots at time $t_1$, $t_2$, $t_3$, and (b), (d), (e) the maximum typicality plots of four points (in green, red, cyan, magenta color) at time $t_1$, $t_2$, $t_3$

In Fig. 6 (a) and (b), the streaming data just started at time $t_1$ and the cyan and magenta color points have low maximum typicality value because there are no prototypes close to them. In Fig. 6 (c) and (d), the streaming data formed a new class around the cyan color point at time $t_2$. StreamSoNG detected this new class and generated new prototypes for this new class so that the maximum typicality of the cyan color point increased at time $t_2$. In Fig. 6 (e) and (f), the streaming data formed another new class around the magenta color point at time $t_3$. Similarly, StreamSoNG detected this new class as well and generated new prototypes for this class so that the maximum typicality of the magenta color point increased at time $t_3$. A similar analysis on the dataset 2-4 is included in the supplemental material.

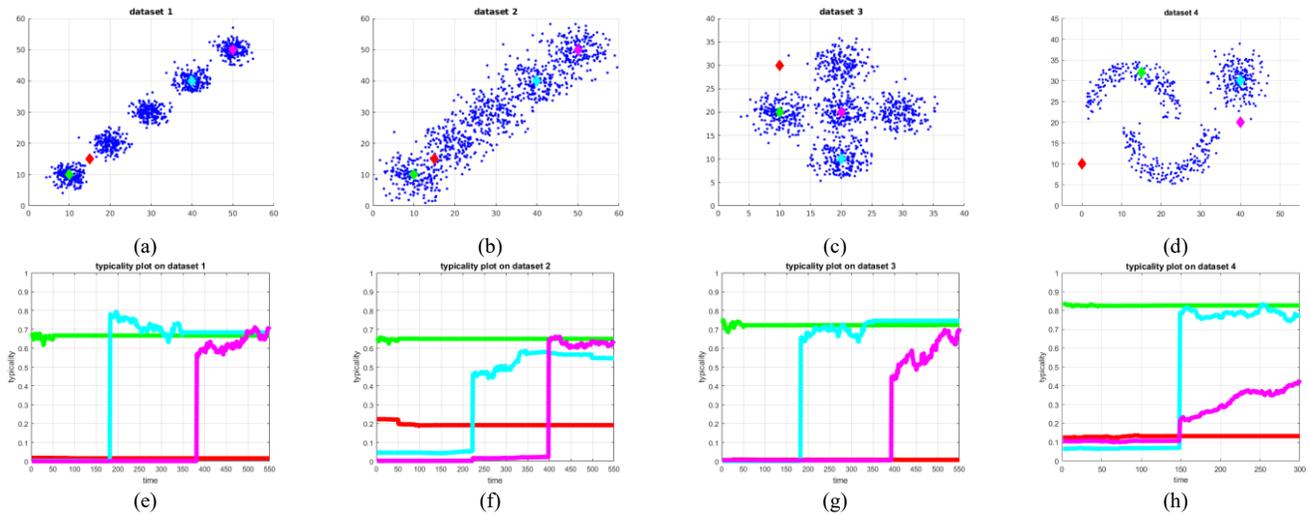
Fig. 7. Plot of maximum typicality for four locations (colored dots) over time on the four synthetic datasets.

Fig. 7 shows the final maximum typicality plot of the four points in the four datasets. In Fig. 7 (a) – (d), 4 diamond symbols (in green, red, cyan, magenta color) in each dataset are studied. Their maximum typicality value plots with respect to time are shown in Fig. 7 (e) – (h).

As we see in Fig. 7, the green diamond symbol always has a high maximum typicality value as the model updates with streaming data because it is always in the middle of a class. The red diamond symbol has a low typicality all the time because it is always in the sparse area. The cyan diamond symbol and the magenta diamond symbol are two interesting cases. Their typicality values are low at the beginning, then become high as more streaming data form a new cluster around their regions and the algorithm creates a new class around them.

### E. Experiment 4: Test on a real-world texture image dataset

In this experiment, we run the StreamSoNG algorithm on the UMD texture dataset [43]. We use 400 images for initialization that includes pebbles and bricks images. We use another 600 images as the streaming set that includes pebbles, bricks and plaid types of images. The plaid type image is new in the streaming set that is not included in the initialization set. Three examples of each type of image are shown in Fig. 8. This is a surrogate test for the intended use of determining typicalities of seafloor textures displayed in SAS imagery.

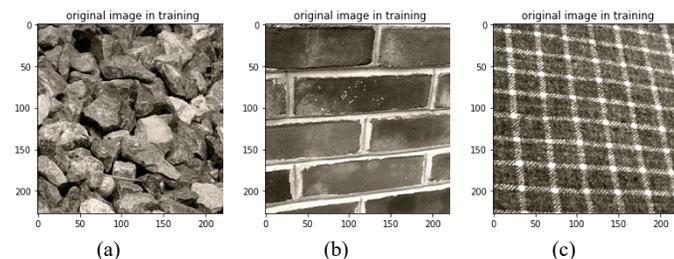
Fig. 8. Three examples of each type of images: (a) Pebbles (b) Bricks (c) Plaid

First, we generate a texture feature vector from a deep pre-trained Convolutional Neural Network, Resnet18 [44] with classification layers removed to extract 512 features from these images. Then we train an autoencoder model with three hidden layers on the 512 features of the initialization set. The Resnet18 pre-trained model and encoder in the autoencoder are used together to process the streaming images to get 16 features. Then we run the StreamSoNG algorithm on the extracted 16 features. StreamSoNG achieves 81.3% precision on the entire streaming set and 95.7% precision on the streaming samples that have maximum typicality value higher than 0.2. StreamSoNG detects the plaid class in the streaming set and produces a new class label for the plaid type of image. Fig. 9 shows three examples of typicality changes in each class.

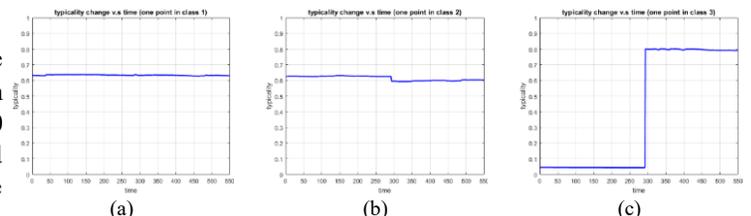
Fig. 9. Typicality value plot for (a) a streaming sample from the pebbles class, (b) a streaming sample from the bricks class, (c) a streaming sample from the new plaid class

The first two samples in the pebbles and bricks class consistently have high typicality values as streaming data is fitted into the model. The typicality value of the third example from the new plaid class has low typicality value at the beginning and high typicality value when a new plaid class is created in the model.

One application of our StreamSoNG model is to detect the environment using drones. Fig. 10 mimics a scenario that a drone flies from a brick region to a pebble region. In Fig. 10 (a), the drone is completely in the brick region. It gradually flies over to the pebble region as Fig. 10 (b) - (j) show. In the end, the drone is completely in the pebble region as Fig. 10 (k) shows.

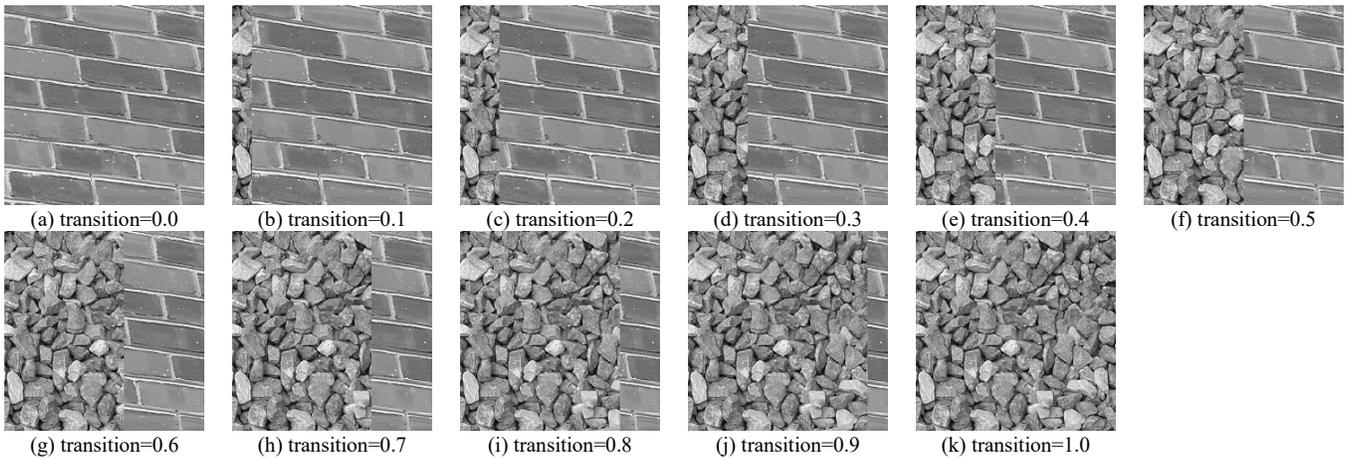
Fig. 10. A sequence of transition images from the brick region (class 2) to the pebble region (class 1)

We keep track of the typicalities of the sequence of transition images in Fig. 10 while running the StreamSoNG model. The typicalities of the images to the pebble class (class 1) and the brick class (class 2) are shown in Fig. 11.

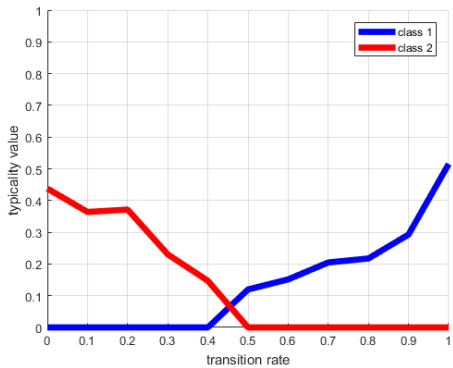
Fig. 11. Typicalities of the sequence of transition images to the pebble class (class 1) and the brick class (class 2)

As the environment shifts from the brick region (class 2) to the pebble region (class 1), the typicality value to class 1 is increasing and the typicality value to class 2 is decreasing. Our StreamSoNG reflects the environment transition fact in the typicality plot successfully.

## VI. CONCLUSION

In this paper, we proposed a soft streaming classification algorithm. This is particularly useful for situations where the streaming data classes are overlapped, for example classifying land cover from drone imagery where individual images may contain more than one class and where classes blend from one to another. Each class, both during initialization and in the new structure discovery module, is summarized via a set of Neural Gas prototypes that are then used in a possibilistic K-nearest neighbor algorithm to assign typicalities to each incoming point. Class footprints (the NG prototypes) are incrementally updated. StreamSoNG's performance is excellent on both synthetic and real datasets both from a precision standpoint after hardening the possibilistic labels, and from the standpoint of the actual possibilistic labels assigned to the incoming streaming data. There are several avenues for investigation within the actual structure of StreamSoNG including choices of parameters, scaling of typicalities, assigning and updating fuzzy class memberships of prototypes, and varying the number of prototypes per class, using for example Growing Neural Gas. We intend to couple this steaming classification approach to the problem of environmentally aware classifier fusion. For example, the typicalities can be used to build a fuzzy measure that drives a Choquet integral fusion of a series of deep nets trained on specific environments. A parallel problem that we will investigate is how to generate a new classifier when a novel environment is discovered.